%% file: sample-sigconf.tex
\documentclass[sigconf]{acmart}
\AtBeginDocument{%
  }

\copyrightyear{2025}
\acmYear{2025}
\setcopyright{acmlicensed}
\acmConference[KDD '25] {Proceedings of the 31st ACM SIGKDD Conference on Knowledge Discovery and Data Mining V.2}{August 3--7, 2025}{Toronto, ON, Canada.}
\acmBooktitle{Proceedings of the 31st ACM SIGKDD Conference on Knowledge Discovery and Data Mining V.2 (KDD '25), August 3--7, 2025, Toronto, ON, Canada}
\acmISBN{979-8-4007-1454-2/25/08}
\acmDOI{10.1145/3711896.3737432}

\usepackage[utf8]{inputenc}

\usepackage{microtype}

\usepackage{graphicx}
\usepackage{booktabs}
\usepackage{multicol}
\usepackage{multirow}
\usepackage{siunitx}    
\usepackage{array}      
\usepackage{xcolor} 
\usepackage{amsmath}
\usepackage{comment}
\usepackage{ltablex}
\usepackage{subcaption}
\usepackage{lscape}   
\usepackage{pdflscape} 
\usepackage{longtable} 
\usepackage{array}     
\usepackage{ltablex}   
\usepackage{booktabs}  
\usepackage{xspace} 
\usepackage{arydshln}
\usepackage{subcaption} 
\usepackage{caption}

\definecolor{lu}{RGB}{188, 0, 0} 

\newcommand\fancynameB{$\Delta$\textit{SePer}\xspace}

\newcommand{\fancyname}{\ensuremath{\textit{SePer}}\xspace}

\begin{document}

\title{ScIRGen: Synthesize Realistic and Large-Scale RAG Dataset for Scientific Research}

\author{Junyong Lin}
\authornote{Both authors contributed equally to this research.}
\affiliation{%
    \institution{The Hong Kong University of Science and Technology (Guangzhou)}
    \city{Guangzhou}
  \state{Guangdong}
  \country{China}
}
\email{jlin159@connect.hkust-gz.edu.cn}

\author{Lu Dai}
\authornotemark[1]
\affiliation{%
    \institution{The Hong Kong University of Science and Technology (Guangzhou)}
  \city{Guangzhou}
  \state{Guangdong}
  \country{China}
}
\affiliation{
    \institution{The Hong Kong University of Science and Technology}
    \city{Hong Kong SAR}
    \country{Hong Kong}
}
\email{ldaiae@connect.ust.hk}

\author{Ruiqian Han}
\affiliation{%
    \institution{The Hong Kong University of Science and Technology (Guangzhou)}
  \city{Guangzhou}
  \state{Guangdong}
  \country{China}
}
\email{rhan464@connect.hkust-gz.edu.cn}

\author{Yijie Sui}
\affiliation{
  \institution{Institute of Tibetan Plateau Research, Chinese Academy of Sciences}
  \city{Beijing}
  \country{China}
}

\author{Ruilin Wang}
  
\author{Xingliang Sun}
\affiliation{%
  \institution{Lanzhou University}
  \city{Lanzhou}
  \state{Gansu}
  \country{China}}

\author{Qinglin Wu}
\affiliation{%
  \institution{Institute of Tibetan Plateau Research, Chinese Academy of Sciences}
  \city{Beijing}
  \country{China}}

\author{Min Feng}
\affiliation{%
  \institution{Institute of Tibetan Plateau Research, Chinese Academy of Sciences}
  \city{Beijing}
  \country{China}
  }
\affiliation{%
  \institution{College of Resources and Environment, University of Chinese Academy of Sciences}
  \city{Beijing}
  \country{China}
  }

\author{Hao Liu}
\authornote{Correspondence author.}
\affiliation{
    \institution{The Hong Kong University of Science and Technology (Guangzhou)}
    \city{Guangzhou}
    \state{Guangdong}
    \country{China}
}
\affiliation{
    \institution{The Hong Kong University of Science and Technology}
    \city{Hong Kong SAR}
    \country{Hong Kong}
}
\email{liuh@ust.hk}

\author{Hui Xiong}
\authornotemark[2]
\affiliation{
    \institution{The Hong Kong University of Science and Technology (Guangzhou)}
    \city{Guangzhou}
    \state{Guangdong}
    \country{China}
}
\affiliation{
    \institution{The Hong Kong University of Science and Technology}
    \city{Hong Kong SAR}
    \country{Hong Kong}
}
\email{xionghui@ust.hk}

\renewcommand{\shortauthors}{Junyong Lin et al.}

\begin{abstract}
  Scientific researchers need intensive information about datasets to effectively evaluate and develop theories and methodologies. The information needs regarding datasets are implicitly embedded in particular research tasks, rather than explicitly expressed in search queries. However, existing scientific retrieval and question-answering (QA) datasets typically address straightforward questions, which do not align with the distribution of real-world research inquiries. To bridge this gap, we developed ScIRGen, a dataset generation framework for scientific QA \& retrieval that more accurately reflects the information needs of professional science researchers, and uses it to create a large-scale scientific retrieval-augmented generation (RAG) dataset with realistic queries, datasets and papers. Technically, we designed a dataset-oriented information extraction method that leverages academic papers to augment the dataset representation.  We then proposed a question generation framework by employing cognitive taxonomy to ensure the quality of synthesized questions. We also design a method to automatically filter synthetic answers based on the perplexity shift of LLMs, which is highly aligned with human judgment of answers' validity. Collectively, these methodologies culminated in the creation of the 61k QA dataset, ScIRGen-Geo. We benchmarked representative methods on the ScIRGen-Geo dataset for their question-answering and retrieval capabilities, finding out that current methods still suffer from reasoning from complex questions. This work advances the development of more sophisticated tools to support the intricate information needs of the scientific community.
\end{abstract}

\ccsdesc[500]{Information systems~Retrieval tasks and goals}
\ccsdesc[500]{Computing methodologies~Natural language processing}
\ccsdesc[300]{Applied computing~Earth and atmosphseric sciences}

\keywords{Retrieval-Augmented Generation (RAG), Large Language Models (LLMs), Scientific Workflows, Dataset Generation, Synthetic Dataset, Information Retrieval, Question Answering, Geoscience}


\maketitle
\input{subtex/introduction}
\input{subtex/dataset_collection}
\input{subtex/dataset_analysis}

\input{subtex/experiments}
\input{subtex/relatedwork}
\input{subtex/limitation_conclusion}

\bibliographystyle{ACM-Reference-Format}
\bibliography{sample-sigconf}

\appendix

\input{subtex/prompt_appendix}

\end{document}

%% file: subtex/introduction.tex
\section{Introduction}
\begin{figure}
    \centering
    \includegraphics[width=1\linewidth]{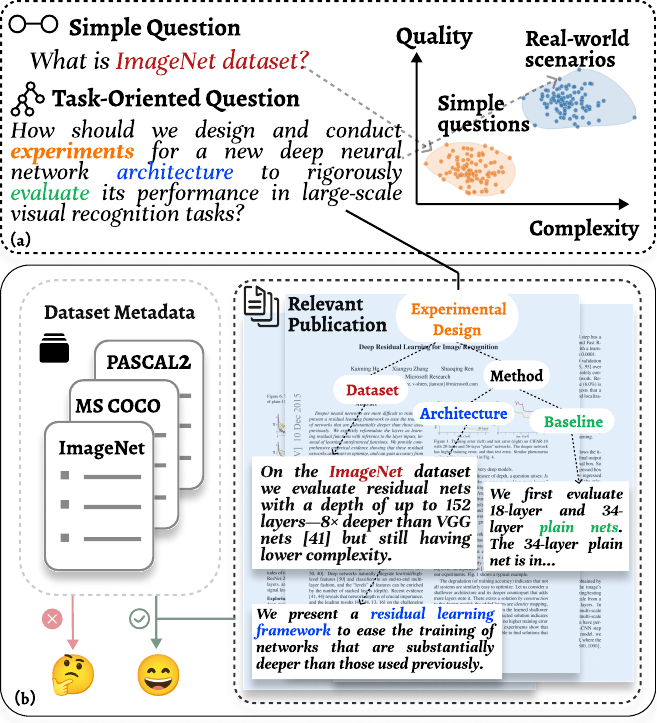}
    \caption{Traditional queries fall short in task-oriented scenarios, where queries are sophisticated and typically found in academic literature rather than dataset metadata.}
    \label{fig:teaser}
    \vspace{-0.6cm}
\end{figure}

Effective Decision-making in scientific research relies on reasoning through comprehensive information and available scientific resources~\cite{specter:2020, datafinder:2023}. Recently, large language models (LLMs) have shown remarkable capabilities for answering open-domain questions, and several have been fine-tuned for scientific applications~\cite{scibert:2019, galactica:2022, forge:2023, k2:2024}. However, they suffer from problems such as hallucinations and difficulty in source attribution. 

Retrieval-Augmented Generation (RAG) frameworks address these issues by integrating two components: 1) a retriever that retrieves relevant information through lexical and semantic matching from a database, and 2) a generator that incorporates retrieved evidence to generate reliable and traceable responses~\cite{rag1:2020, rag2:2020}, thereby leveraging outside information to enhance factual correctness and verifiability. 
However, two critical gaps persist when aligning these systems with the realities of scientific research.  

First, the queries in many existing RAG datasets are overly simplistic~\cite{nq:2019, bright:2025} (Figure \ref{fig:teaser}(a)). In contrast, real research workflows involve complex, task-oriented questions that emerge at each stage of problem-solving, such as data collection, analysis, experimental design, and evaluation~\cite{sciagentbench:2025}. For example, a researcher in environmental science might ask, \textit{“What methods are used for processing eddy covariance data to ensure quality and accuracy in arid regions?”} during data collection, and later inquire, \textit{“What specific features should be considered in the quality evaluation of hydrometeorological flux data?”} during evaluation. Unlike direct resource-seeking requests (e.g., “Find the MNIST dataset”), these queries are task-oriented, reflecting more nuanced information needs throughout the research workflow. This discrepancy underscores a distribution gap between authentic research queries and the straightforward questions typically found in existing retrieval datasets~\cite{litsearch:2024}.


Second, many scientific datasets come with only basic metadata (Figure \ref{fig:teaser}(b)), which lacks the detailed context needed to address complex research questions~\cite{datafinder:2023}. Specifically, information such as background, motivation, challenges, and experimental outcomes is often found in relevant papers, which can support researchers in interpreting existing work and guiding future investigations. These supplementary materials are often overlooked, which limits the utility of current data-driven research tools.

Prior works have explored building datasets with more complex questions and verifiable evidence through methods such as crowdsourcing \cite{litsearch:2024}, question composition \cite{hotpotqa:2018, musique:2022}, mining from public forums \cite{bright:2025}, and LLM-based generation \cite{llm_data_gen:2024}. However, collecting realistic data in specialized domains can be prohibitively expensive, and in many fields, publicly available datasets are either unreliable or simply do not exist. Meanwhile, synthetic data often raises concerns about quality. As a result, realistic RAG datasets for scientific research remain scarce. This gap raises a key question: \textbf{Is there a reliable and convenient way to construct RAG datasets that capture high-quality scientific queries by design?}

To this end, we propose ScIRGen, a systematic framework for constructing realistic retrieval and QA datasets for scientific research. For context collection, we organize resources in a dataset-oriented manner and employ an information extraction pipeline to gather and parse relevant research papers. This not only grounds information in real resources but also enhances the depth of their representations. For question generation, we leverage Graesser and Pearson’s \cite{graesserquestion:1994} question scheme—an extensive, pedagogically grounded taxonomy that delineates various question types and prototypes, which enables large language models to produce controllable and purposeful questions. Our analysis shows that the questions generated under this scheme span a broad spectrum of cognitive complexity, ensuring both diversity and challenge. For answer collecting, we introduce a novel statistical filter to reliably select valid answers generated by large language models. Experiments show that this filter is highly aligned with human judgment of answers' validity. We use this methodology to create \textbf{ScIRGen-Geo}, a new dataset for retrieval and question answering in the geoscience domain. \textbf{ScIRGen-Geo} consists of 61K QA spanning all 18 question types constructed from Graesser and Pearson’s taxonomy, alongside a corpus comprising 3,345 datasets and relevant papers. It supports a range of tasks, such as dataset retrieval and question answering, and is particularly designed to address the complex information demands in scientific discovery workflows in geoscience. 
\begin{figure*}[h]
    \centering
    \includegraphics[width=\textwidth]{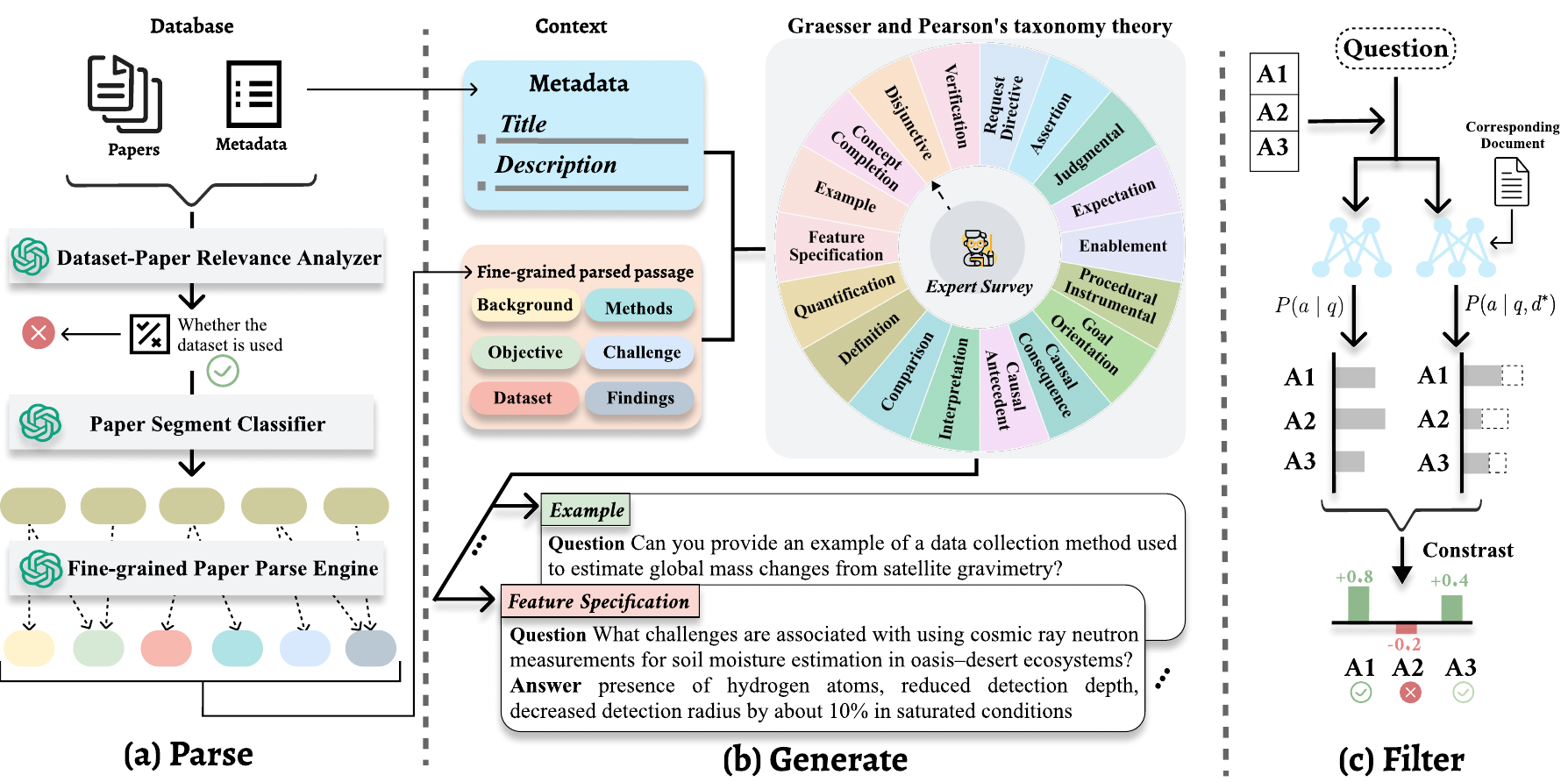}
    \caption{The QA data generation pipeline. (a) Gather dataset metadata along with a corpus of relevant papers, from which we extract relevant passages. (b) Guided by Graesser and Pearson's taxonomy theory, we use the extracted metadata and passages as context to generate QA data. (c) We employ the innovative $\Delta \fancyname$ method to filter the generated QA data.}
    \vspace{-10pt}
    \label{fig:pipeline}
\end{figure*}
To summarize, our contributions are three-fold:
\begin{enumerate}
    \item \textbf{Novel Framework}: A systematic, end-to-end approach that integrates context representation, question generation, and answer filtering to build realistic retrieval-augmented QA datasets.
    \item \textbf{New Dataset}\footnote{https://github.com/ScIRGen/ScIRGen}: a large-scale geoscience dataset containing diverse, research-oriented queries and grounded contexts, developed under our framework.

    \item \textbf{Empirical Validation}: In-depth evaluation and analysis of the framework and dataset, with a strong benchmark for scientific retrieval and question-answering tasks.
\end{enumerate}

%% file: subtex/dataset_collection.tex
\section{Dataset Collection}

This section presents the ScIRGen framework, which (Section 2.1) harvests domain-specific datasets and their linked publications, (Section 2.2) synthesizes diverse, task-oriented QA pairs, and (Section 2.3) performs automatic quality filtering (Section 2.3)—steps that collectively produce the ScIRGen-Geo dataset, a reliable, context-rich corpus for advanced scientific QA and retrieval.


\subsection{Corpus Curation}

To construct a high-quality scientific corpus with rigorous dataset-paper alignment, we designed a multi-stage curation workflow (Figure \ref{fig:pipeline}(a)) combining systematic dataset collection, relevance filtering, and hierarchical content parsing.

\subsubsection{Paper-Dataset Matching}

We initiated our workflow by curating all 6,819 datasets from the National Tibetan Plateau/Third Pole Environment Data Center\footnote{https://data.tpdc.ac.cn/home}, each accompanied by human-written metadata (titles, dataset description, research topics). These datasets were systematically mapped to 6,755 domain-specific publications, each of which directly refers to or contains relevant information related to this dataset.

We propose a Dataset-Paper Relevance Analyzer based on LLMs that leverages semantic matching to identify papers that substantively utilize specific datasets. Beyond simple citation matching, we prompt LLMs to analyze methodology alignments and experimental configurations that indicate dataset usage. The matching process integrates dataset metadata, technical specifications, and domain attributes to evaluate both explicit references and implicit indicators. For each dataset-paper pair, we generate a structured assessment examining methodology alignment, terminology usage, and results consistency, enabling robust identification of relevant papers even in cases of ambiguous citations. 

\definecolor{questionword}{RGB}{0,102,204}  
\definecolor{keyword}{RGB}{178,34,34}       

\begin{table*}[htbp]
    \centering
    \scriptsize
    \resizebox{\linewidth}{!}{%
    \begin{tabular}{p{0.13\linewidth} p{0.40\linewidth} p{0.43\linewidth}}
        \toprule
        \textbf{Question Category} & \textbf{Explanation} & \textbf{Example} \\
        \midrule
        \multicolumn{3}{l}{\textbf{Short answer}} \\
        \midrule
        \textbf{Verification} & Verification questions seek a simple 'yes' or 'no' answer to confirm specific details. 
        & \textbf{\textcolor{questionword}{Can}} ground temperature reflect the influence of \textcolor{keyword}{water in permafrost} on \textcolor{keyword}{thermokarst lakes}? \\
        
        \textbf{Disjunctive} & Disjunctive questions present multiple options, asking the researcher to identify which one is applicable.
        & \textcolor{questionword}{\textbf{Which}} index is more effective for extracting \textcolor{keyword}{glacier lakes} (MNDWI, NDWI, AWEI, or other indexes)? \\
        
        \textbf{Concept completion} & Concept completion questions start with 'Who?', 'What?', 'When?', or 'Where?' to prompt the identification or completion of a specific term or defined element. 
        & \textcolor{questionword}{\textbf{Where}} do they typically develop? \textcolor{questionword}{\textbf{When}} do they form? \\
        
        \textbf{Feature specification} & Feature specification questions inquire about the properties or characteristics of a concept, object, or phenomenon. 
        & \textbf{\textcolor{questionword}{What}} are the \textbf{\textcolor{questionword}{characteristics}} of \textcolor{keyword}{reservoir changes}? \\
        
        \textbf{Quantification} & Quantification questions seek numerical or measurable information. 
        & \textbf{\textcolor{questionword}{How many}} \textcolor{keyword}{thermokarst lakes} are there on the \textcolor{keyword}{Tibetan Plateau}? \\
        
        \midrule
        \multicolumn{3}{l}{\textbf{Long answer}} \\
        \midrule
        \textbf{Definition} & Definition questions ask researchers to explain the meaning of a specific term or concept.
        & \textbf{\textcolor{questionword}{What is}} \textcolor{keyword}{thermokarst lake drainage}? \\
        
        \textbf{Example} & Example questions ask for instances that illustrate a particular scientific concept.
        & Can you provide an \textbf{\textcolor{questionword}{example}} of a \textcolor{keyword}{thermokarst lake} in the \textcolor{keyword}{Tibetan Plateau} rapidly draining? \\
        
        \textbf{Comparison} & Comparison questions require researchers to identify similarities and/or differences between two or more scientific resources or concepts.
        & What are the \textbf{\textcolor{questionword}{similarities and differences}} between \textcolor{keyword}{thermokarst lakes} and \textcolor{keyword}{small lakes} in permafrost regions? \\
        
        \textbf{Interpretation} & Interpretation questions ask researchers to infer underlying rules of their observed data patterns.
        & \textbf{\textcolor{questionword}{What}} physical phenomena can be \textbf{\textcolor{questionword}{inferred from}} \textcolor{keyword}{static or periodic patterns} in \textcolor{keyword}{vibration data}? \\
        
        \textbf{Causal Antecedent} & Causal antecedent questions inquire about the reasons or causes behind an event, trend. 
        & \textbf{\textcolor{questionword}{Why}} does the \textcolor{keyword}{Tibetan Plateau} influence \textcolor{keyword}{global climate}? \\
        
        \textbf{Causal Consequence} & Causal consequence questions ask about the outcomes or results that follow from a specific event, trend.
        & \textbf{\textcolor{questionword}{What happens after}} \textcolor{keyword}{thermokarst lakes expand}? \\
        
        \textbf{Goal Orientation} & Goal orientation questions investigate the objectives or intentions behind the creation of a dataset, publication, or research project.
        & \textbf{\textcolor{questionword}{Why}} use \textcolor{keyword}{multi-temporal data fusion} \textbf{\textcolor{questionword}{to}} monitor \textcolor{keyword}{land cover types} in mountainous areas? \\
        
        \textbf{Instrumental or Procedural} & Instrumental or procedural questions ask how to achieve certain goals. 
        & What is the \textbf{\textcolor{questionword}{detailed process}} of \textcolor{keyword}{thermokarst lake extraction}? \\
        
        \textbf{Enablement} & Enablement questions focus on identifying the resources or conditions that enable an agent to perform a specific action.
        & \textbf{\textcolor{questionword}{What}} technological advancements \textbf{\textcolor{questionword}{enabled}} the collection of \textcolor{keyword}{high-resolution climate data}? \\
        
        \textbf{Expectation} & Expectation questions inquire about anticipated outcomes or reasons why expected results did not occur.
        & \textbf{\textcolor{questionword}{Why}} are \textcolor{keyword}{small glacial lakes} \textbf{\textcolor{questionword}{rarely}} reported as having \textcolor{keyword}{outburst events}? \\
        
        \textbf{Judgmental} & Judgmental questions ask researchers to express their opinions or evaluations.
        & \textbf{\textcolor{questionword}{What do you think}} of the future risk of \textcolor{keyword}{small glacial lake outbursts}? \\
        
        \textbf{Assertion} & Makes a statement indicating a lack of knowledge or does not understand an idea.
        & \textbf{\textcolor{questionword}{I don't understand}} why the risk of \textcolor{keyword}{glacial lake outbursts} is high? \\
        
        \textbf{Request/Directive} & Request or directive questions involve asking researchers to perform specific tasks, such as summarizing information, analyzing data, or conducting searches.
        & \textbf{\textcolor{questionword}{Please optimize}} the \textcolor{keyword}{glacial lake outburst risk assessment model}. \\
        
        \bottomrule
    \end{tabular}%
    }
    \caption{Question Types in Graesser, Person, and Huber (1992) with Definitions and Examples. Blue highlights the question prompt words that frame the inquiry, while dark red signifies the technical keywords anchoring the specialized content.}
     \label{tab:graesser-scheme}
     \vspace{-20pt}
\end{table*}

\subsubsection{Multi-Aspect Parsing}
For matched papers, we implement a systematic parsing framework to construct comprehensive research profiles to ensure precision and utility.

For papers that are relevant to a dataset, we design a two-stage content extraction methodology. In the first stage, we structurally partition each paper into five sections: abstract \& introduction, related work, method, experimental analysis, and conclusion. For each section, we design prompts to enable the LLM to semantically extract six categories of dataset-related content: background, research objectives, methods, challenges, dataset usage, and findings. To further ensure extraction accuracy and mitigate the impact of LLM hallucinations, the second stage prompts the LLM to evaluate each extracted content against the dataset metadata, filtering out irrelevant content that may have been erroneously extracted, thereby mitigating the adverse effects of model hallucinations.

The two-stage extraction approach ensures the comprehensive curation of all dataset-related information within papers, with the extracted content capturing dataset-related information at both structural and semantic levels, thereby enabling sophisticated understanding of complex scientific queries.

The curation workflow ultimately yielded 3,345 dataset metadata entries paired with 943 high-relevance papers (Table \ref{tab:dataset_stats}(b)) and 15,389 extracted paragraph units spanning six distinct aspects (Table \ref{tab:dataset_stats}(c)), providing a professional and comprehensive graph-structured data foundation for QA data construction in next step.
\vspace{-20pt}

\subsection{Question-Answering Data Generation}

Our Question-Answering (QA) data generation framework (Figure \ref{fig:pipeline}(b)) employs Graesser and Pearson’s taxonomy \cite{graesserquestion:1994}, a theoretically grounded schema comprising 18 distinct question types for categorizing inquiry patterns in a pedagogical taxonomy (Table \ref{tab:graesser-scheme}). This taxonomy differentiates between short-answer (e.g., verification, quantification) and long-answer (e.g., causal consequence, instrumental procedural) questions, systematically addressing both factual questions and complex reasoning questions. We adopted this framework due to three key advantages: \textbf{1)} its cognitive structure aligns well with professional scientific research question patterns; \textbf{2)} its comprehensiveness supports multifaceted exploration of datasets; and \textbf{3)} its pedagogical rigor ensures that the generated questions probe both foundational knowledge and analytical depth.

For datasets with associated publications, we utilize the taxonomy for multi-type generation, producing three QA pairs per question type, resulting in 54 pairs per dataset (18 types × 3 pairs/type). This process is further enhanced by integrating human expertise: \textbf{1)} Domain specialists refine Graesser’s original definitions to incorporate geoscience-specific knowledge and provide annotated examples (Table \ref{tab:graesser-scheme}). \textbf{2)} An in-context learning prompting method is employed, utilizing these refined definitions and examples while adhering to strict output constraints, such as prioritizing open-ended questions or focusing on methodological inquiries (Table \ref{prompt:main_generation}).

For datasets lacking associated publications, an additional preprocessing stage is introduced (Table \ref{prompt:generate_for_without_pdf}) before generating QA pairs using our primary method. This stage involves analyzing metadata (titles and descriptions) to identify the eight most likely question types for a given dataset. For instance, 'Quantification' questions are prioritized for datasets featuring temporal coverage metrics. This tailored approach ensures the effective generation of high-quality and relevant QA data, even in the absence of a full publication context. Consequently, each such dataset is paired with eight QA pairs.

The framework has been applied to generate 61,376 QA pairs across a broad spectrum of question types (Table \ref{tab:dataset_stats}(a)), addressing diverse utility requirements. For concise inquiries, it encompasses categories such as ‘Verification,’ ‘Disjunctive,’ and ‘Concept Completion.’ For more in-depth explorations, it addresses categories including ‘Example,’ ‘Definition,’ and ‘Comparison.’ By systematically aligning these question categories with authentic research inquiries, the framework effectively captures the multifaceted information needs of researchers.

\subsection{Data Filtering}

\begin{table}[h]

\begin{tabular}{@{}lllll@{}}
\toprule
                           &        & \textbf{Precision} & \textbf{Recall} & \textbf{F1}   \\
\midrule
\textit{w/o Filter}                 &      & 0.86      & -      & -    \\
\midrule
\multirow{3}{*}{\textit{w/ Filter}} & GPT-3.5-turbo  & 0.90      & 0.43   & 0.59 \\
\cmidrule(l){2-5} 
 & GPT-4  & 0.89      & 0.49   & 0.64 \\
\cmidrule(l){2-5} 
                & \fancynameB & \textbf{0.95}      & \textbf{0.71}   & \textbf{0.81} \\
\bottomrule
\end{tabular}
\caption{Comparison of answer filter performance.}
\label{tab:seper}
\vspace{-0.6cm}
\end{table}

Filter is not a trivial task~\cite{filter_roundtrip:2019}. While the correctness of problems of mathematical and logic reasoning can be verified by numerical equivalence~\cite{mathverify:2024, r1:2025}, automatically verifying the quality of free-form natural language generation lacks a clear and reliable methodology.

\begin{figure}[htph]
    \centering
    \begin{subfigure}[b]{0.45\linewidth}
        \centering
        \includegraphics[width=\linewidth]{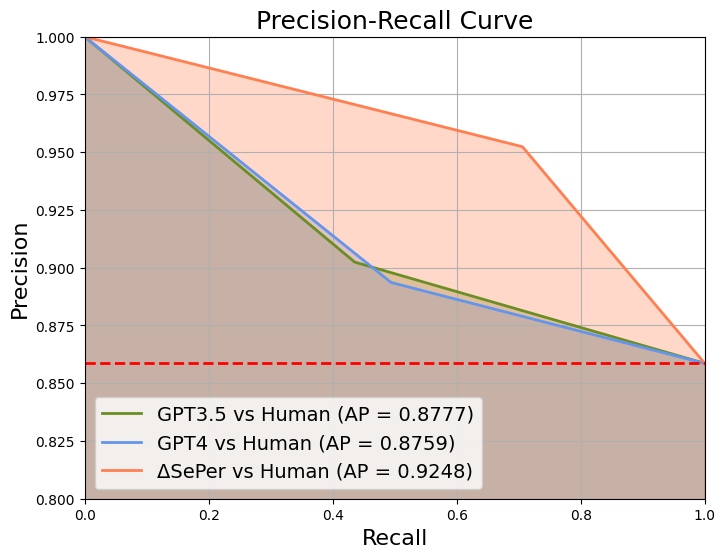}
        \caption{Precision-Recall Curve}
        \label{fig:pr_curve}
    \end{subfigure}
    \hfill
    \begin{subfigure}[b]{0.45\linewidth}
        \centering
        \includegraphics[width=\linewidth]{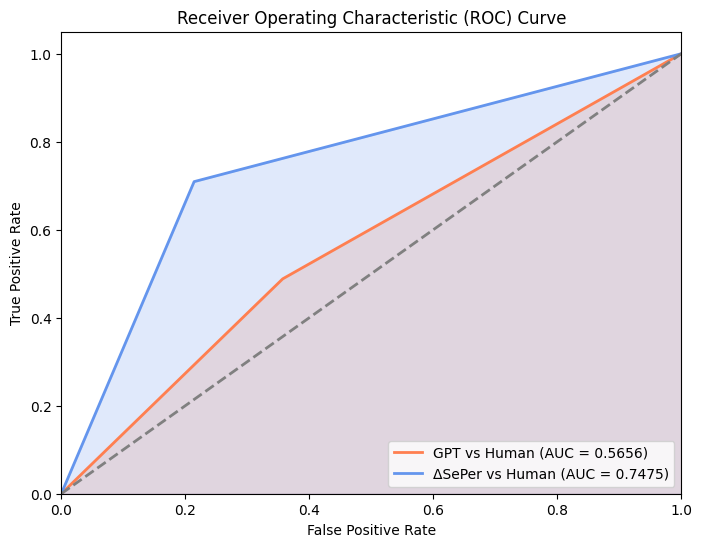}
        \caption{ROC Curve}
        \label{fig:roc_curve}
    \end{subfigure}
    
    \caption{Answer Filter Assessment: \emph{\fancyname-Filter} vs. GPT-3.5 and GPT-4}
    \vspace{-0.5cm}
    \label{fig:pr_roc}
\end{figure}

In this section, we rigorously define our automatic QA data filtering method (Figure \ref{fig:pipeline}(c)), \emph{\fancyname-Filter}, which is an adaptation of the \fancyname framework~\cite{seper:2025}. In its original form, \fancyname was designed to evaluate the utility of retrieved content in retrieval-augmented generation (RAG) systems. Here, we repurpose this idea to assess the faithfulness of generated answers with respect to a provided context.

Let $q \in \mathcal{Q}$ denote a question, $d \in \mathcal{D}$ denote the associated context (or document), and $a^* \in \mathcal{A}$ denote the ground truth answer. Given a language generator $G$, we define the conditional probability $P_M(a^* \mid q)$
as the confidence of $G$ in generating the answer $a^*$ when provided only with the question $q$, and $P_M(a^* \mid q, d)$ as the confidence provided with both the question $q$ and the context $d$.
Then \fancynameB is used to quantify the impact (or \emph{belief shift}) that the context $d$ has on the generation of the answer $a^*$, which is formally defined as:
\begin{equation}
\Delta \fancyname = P_M(a^* \mid q, d) - P_M(a^* \mid q).
\end{equation}

Different from \fancynameB, \text{SePerFilter} is given $q$ and $d^*$ and to assess the quality of the answer $a$. The underlying hypothesis is that if the answer $a^*$ is both correct and faithful to the context $d$, then incorporating $d$ should increase the generator's confidence in $a^*$, leading to \fancynameB > 0.
Conversely, if $a^*$ is incorrect or unfaithful to the context, the additional information provided by $d$ will not enhance—and may even diminish—the confidence, resulting in \fancynameB $\leq 0$.

Based on this observation, our filtering criterion for a given question-context-answer triplet $(q,d,a^*)$ is defined as follows:
\begin{equation}
\label{eq:seperfilter}
\text{SePerFilter}(q,d,a^*) = 
\begin{cases}
\text{Accept} & \text{if } \Delta\textit{SePer} > 0, \\
\text{Reject} & \text{if } \Delta\textit{SePer} \leq 0.
\end{cases}
\end{equation}

To assess the validity and robustness of this approach, we tested the filtering alignment with human annotation. 
\paragraph{Dataset Preparation:} We uniformly sub-sampled 100 samples comprising a diverse set of question-context-answer triplets from our full synthetic dataset.
\paragraph{Annotation Protocol:} A group of expert annotators is recruited to independently score each answer by its \emph{Faithfulness}, i.e., whether the answer is supported by and consistent with the context. The human experts are required to rate binarily as Yes or No and write down their reasons.
\paragraph{Metric Computation:} For each triplet, we compute the \fancynameB metric as defined in Equation~\ref{eq:seperfilter}.
\paragraph{Statistical Analysis:} We then measure the filtering accuracy based on the human-annotated scores, and compare \emph{\fancyname-Filter} to strong LLM-as-a-judge methods that leverage GPT-4 to score and filter the answers. As the results shown in Table~\ref{tab:seper} and Figure~\ref{fig:pr_roc}, \emph{\fancyname-Filter} lifts the accuracy of the synthetic dataset by almost 10\% points, hitting a precision of 95\% according to human annotation, which indicates the synthetic dataset as a reliable ground-truth annotation. Notably, for other strong baselines such as GPT-4, while their filtering also increases the cleanliness of the synthetic dataset by 3\% points, it still lags behind our methods both in precision and recall. Moreover, \emph{\fancyname-Filter} does not rely on costly API calls but rather logit output from open-source LLMs, which is also more convenient and cheap for various scenarios.

%% file: subtex/dataset_analysis.tex
\section{Dataset Analysis}

\subsection{Dataset statistics}
\begin{table}[ht]
\centering
\scriptsize

\resizebox{\linewidth}{!}{%
  \begin{tabular}{lrrrr}
    \toprule
    \multicolumn{5}{c}{\textbf{(a) Question Type Distribution}} \\
    \midrule
    \textbf{Question Type} & \textbf{Count} & \textbf{Percentage} & \textbf{Avg. Q. Words} & \textbf{Avg. A. Words} \\
    \midrule
    Verification             & 2030  & 3.31\%   & 18.36 & 1.00 \\
    Disjunctive              & 1697  & 2.76\%   & 21.77 & 4.04 \\
    Concept Completion       & 3386  & 5.52\%   & 15.88 & 6.38 \\
    Feature Specification    & 4380  & 7.14\%   & 18.19 & 22.77 \\
    Quantification           & 4546  & 7.41\%   & 18.47 & 2.81 \\
    \midrule
    \textbf{Short Answer}    & 16039 & 26.13\% & 18.18 & 8.91 \\
    \midrule
    Definition               & 3949  & 6.43\%   & 14.73 & 59.57 \\
    Example                  & 3655  & 5.96\%   & 20.96 & 49.49 \\
    Comparison               & 2735  & 4.46\%   & 27.17 & 65.17 \\
    Interpretation           & 3207  & 5.23\%   & 23.99 & 60.04 \\
    Causal Antecedent        & 4449  & 7.25\%   & 19.49 & 55.70 \\
    Causal Consequence       & 3717  & 6.06\%   & 23.60 & 59.33 \\
    Goal Orientation         & 4196  & 6.84\%   & 21.63 & 52.53 \\
    Instrumental/Procedural  & 4667  & 7.60\%   & 22.36 & 66.83 \\
    Enablement               & 3649  & 5.95\%   & 19.00 & 48.59 \\
    Expectation              & 3034  & 4.94\%   & 19.85 & 46.02 \\
    Judgmental               & 2607  & 4.25\%   & 22.90 & 56.94 \\
    Assertion                & 2826  & 4.60\%   & 18.99 & 49.31 \\
    Request/Directive        & 2646  & 4.31\%   & 21.79 & 64.54 \\
    \midrule
    \textbf{Long Answer}     & 45337 & 73.87\% & 21.09 & 56.53 \\
    \midrule
    \textbf{Total}           & 61376 & 100\% & 20.33 & 44.09 \\
    \bottomrule
  \end{tabular}
}


\resizebox{\linewidth}{!}{%
  \begin{tabular}{@{}cc@{}}
    \begin{tabular}[t]{lr}
      \toprule
      \multicolumn{2}{c}{\textbf{(b)Dataset–paper Link Counts }} \\ 
      \midrule
      \textbf{Number of Datasets} & \textbf{Number of Relevant Papers} \\
      \midrule
      2049   & 0 \\
      819   & 1  \\
      268   & 2  \\
      188   & 3  \\
      22   & 4  \\
      8  & >= 5   \\
      \bottomrule
    \end{tabular}
    &
    \begin{tabular}[t]{lrr}
      \toprule
      \multicolumn{3}{c}{\textbf{(c) Parsed Paper Aspect Coverage}} \\
      \midrule
      \textbf{Extracted Aspect} & \textbf{Count} & \textbf{Avg.Words} \\
      \midrule
      Background          & 3,547 & 147.98 \\
      Research Objective  & 2,504 & 57.62  \\
      Methods             & 2,549 & 138.14 \\
      Challenges          & 2,588 & 60.35  \\
      Dataset             & 2,409 & 80.72  \\
      Findings            & 1,792 & 101.72 \\
      \bottomrule
    \end{tabular}
  \end{tabular}
}
\caption{Comprehensive statistics for the ScIRGen-Geo corpus. (a) Summarises the mix of question categories in ScIRGen-Geo; (b) reports the number of research papers linked to each dataset; (c) shows the information of six parsed aspects from the paper.}
\label{tab:dataset_stats}
\vspace{-20pt}
\end{table}

\begin{figure*}[h]
    \centering
    \includegraphics[width=\textwidth]{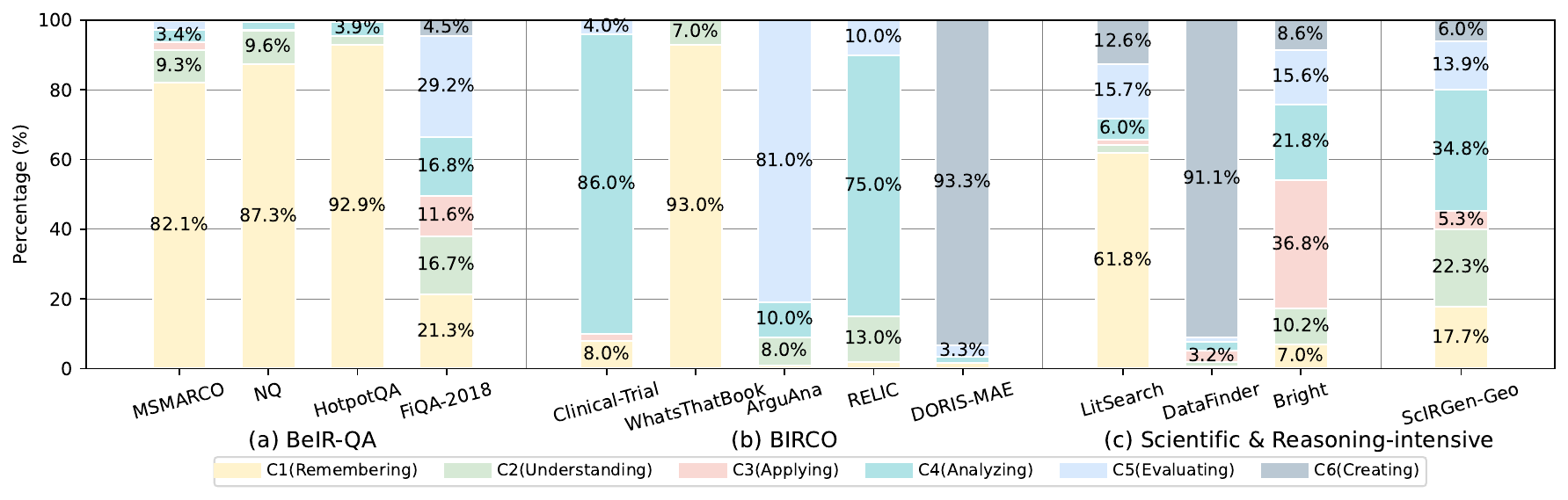}
    \caption{Cognitive Level Distribution Across Different QA Datasets.}
    \label{fig:cognitive_level}
    \vspace{-15pt}
\end{figure*}

Our final dataset comprises 61,376 question-answer (QA) pairs, spanning 18 distinct question types. Detailed statistics, including the count and percentage for each question type, average question length, and average answer length, are presented in Table~\ref{tab:dataset_stats}(a). A notable dichotomy emerges between long-form answers (73.87\% of instances, averaging 56.53 words) and short-form answers (26.13\%, averaging 8.91 words), highlighting distinct requirements for informational depth versus brevity. Instrumental/Procedural (7.60\%) and Causal Antecedent (7.25\%) queries are slightly more prevalent among long-form answers, whereas Feature Specification (7.14\%) and Quantification (7.41\%) queries are more commonly addressed with short-form answers. Lexical patterns reveal extremes in answer complexity: Instrumental/Procedural questions elicit the most elaborate responses (averaging 66.83 words), while Verification questions typically yield binary answers (averaging 1.00 words), underscoring the dataset's linguistic diversity.

As shown in Table~\ref{tab:dataset_stats}(b), 38.9\% of the datasets (1,305 out of 3,354) contain relevant papers. The majority (62.1\%) of these annotated datasets possess one (819 datasets) or two (268 datasets) related papers, while only 0.9\% (22 datasets) are associated with five or more.

Table~\ref{tab:dataset_stats}(c) demonstrates significant variations in aspect coverage. The Background (3,547 instances, averaging 147.98 words) and Methods (2,549 instances, averaging 138.14 words) aspects are richly detailed, providing robust contextual and technical insights for developing QA pairs centered on experimental procedures and foundational research concepts. Additionally, the Dataset aspect mainly describes specific details of a dataset (2,409 instances, averaging 80.72 words), ensuring essential information is presented succinctly. Findings include 1,792 instances (averaging 101.72 words), offering clear overviews of key results. Similarly, Challenges (2,588 instances, averaging 60.35 words) and Research Objectives (2,504 instances, averaging 57.62 words) are frequently and succinctly depicted, efficiently highlighting the critical issues and overarching aims underpinning a dataset. This balanced presentation across various aspects not only caters to diverse research information needs but also facilitates streamlined and effective dataset utilization in scientific research.

\vspace{-5pt}

\subsection{Cognitive Levels Analysis}
\subsubsection{Theoretical Foundation: Revised Bloom's Taxonomy}
The Revised Bloom's Taxonomy (RBT) \cite{revisedBloom’staxonomy:2010} categorizes learning objectives into six progressively complex tiers: Remembering (C1), Understanding (C2), Applying (C3), Analyzing (C4), Evaluating (C5), and Creating (C6). Each tier corresponds to specific cognitive processes, ranging from the basic recall of facts to the generation of novel ideas or solutions. By classifying questions according to the RBT, we can ascertain whether a benchmark primarily assesses surface-level recall or encompasses deeper, more critical thought processes.

Concretely, we identify the dominant mental operation a question demands: whether it requires users merely to recall information (C1), interpret and explain concepts (C2), execute or implement known methods (C3), analyze complex relationships (C4), evaluate ideas against criteria (C5), or create new hypotheses and plans (C6). This emphasis on cognitive processes, rather than topic specificity, enables us to measure how effectively a dataset spans different depths of understanding—a crucial requirement for tasks such as scientific analysis and advanced decision-making.

\subsubsection{Cross-Benchmark Analysis and Differentiation}
We analyze four representative dataset families using the RBT framework:

\textbf{BEIR\cite{beir:2021}} (Figure \ref{fig:cognitive_level}(a)) is a general and comprehensive benchmark for retrieval tasks. For our analysis, we selected test datasets which has question-formed inputs: MSMARCO, NQ, HotpotQA, and FiQA-2018. Questions in MSMARCO and NQ are concentrated in C1 and C2, a distribution that disproportionately favors basic recall and understanding, thereby severely limiting the assessment of higher-order reasoning. HotpotQA’s 97.4\% distribution across C1–C3 reveals its inherent limitations in supporting analytical reasoning, despite its multi-hop design. As FiQA-2018 is directly sampled from real-world user queries, its 648 questions exhibit a more balanced cognitive distribution.

\textbf{BIRCO\cite{birco:2024}} datasets (Figure \ref{fig:cognitive_level}(b)) exhibit highly polarized cognitive distributions that hinder comprehensive evaluation. Overall, these datasets demonstrate significant imbalance: some, such as WhatsThatBook, consist almost entirely of basic tasks (C1–C2), while others, like DORIS-MAE, focus overwhelmingly on high-level creative tasks (C6). Furthermore, datasets like ArguAna are skewed towards higher-level reasoning (C5) without sufficient lower-level scaffolding, while others cluster around mid-level tasks (C4) with minimal variation. This uneven spread restricts the assessment of incremental reasoning across the full cognitive spectrum.

\textbf{Research-Oriented Corpora}, such as LitSearch\cite{litsearch:2024} (Figure \ref{fig:cognitive_level}(c)), tend to skew towards foundational recall (C1), thereby overshadowing deeper analytical tasks. Conversely, DataFinder\cite{datafinder:2023} prioritizes creative tasks (C6) at the expense of foundational cognitive levels. This imbalance may constrain their ability to fully represent the multifaceted reasoning required in scientific discovery. In contrast, BRIGHT\cite{bright:2025}, being directly sampled from the StackExchange community, presents a more balanced distribution of questions.

\textbf{Our Framework} (Figure \ref{fig:cognitive_level}(c)) achieves balanced cognitive coverage through three key design innovations: (1) progressive scaffolding (40.3\% C1–C2 foundational questions enabling knowledge anchoring); (2) targeted emphasis on C4-level analytical reasoning (a 20.4\% weighting, significantly higher than BEIR’s maximum of 3.4\%, for systematic reasoning evaluation); and (3) creative synthesis (9.8\% C6 tasks that incorporate hypothesis generation and cross-domain transfer). Among existing research, both BRIGHT and FiQA demonstrate impressive balance; however, their distributions are derived from real-world community data. In contrast, our framework is unique in achieving a balanced cognitive distribution through synthetic data generation methods.
The cognitive diversity index ($\mathcal{D}=1-\sum_{i=1}^6 p_i^2$) quantifies this advancement: our $\mathcal{D}=0.91$ surpasses BEIR ($0.62 \pm 0.11$) and BIRCO ($0.38 \pm 0.29$), demonstrating a superior capacity to evaluate full-spectrum reasoning abilities. This structured approach addresses the fragmentation observed in existing benchmarks, where 76\% of datasets concentrate more than 70\% of their content within two or fewer cognitive levels.

\begin{figure}[h]
    \centering
    \includegraphics[width=0.5\textwidth]{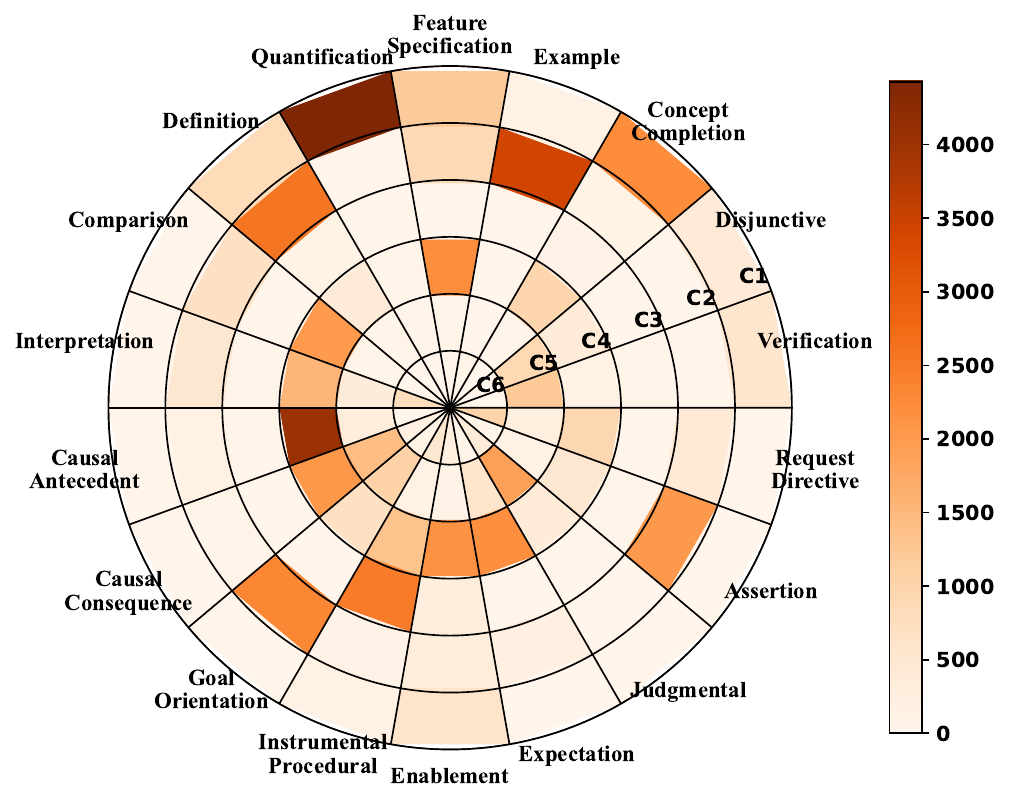}
    \caption{Diagram mapping each Graesser question type in ScIRGen-Geo to revised Bloom's cognitive level and shaded area denotes frequency.}
    \label{fig:type_level}
    \vspace{-22pt}
\end{figure}

Figure~\ref{fig:type_level} further illustrates the distribution of question types across Bloom's six tiers. For instance, C1 (Remembering) is predominantly characterized by tasks such as Quantification (4424) and Concept Completion (2219), reflecting basic fact retrieval. In contrast, C4 (Analyzing) features a significant number of Causal Antecedent (4010) and Causal Consequence (2024) questions, underscoring a focus on dissecting relationships. Similarly, C5 (Evaluating) and C6 (Creating) encompass rich evaluative (e.g., Judgmental, 1891) and generative (e.g., Request/Directive, 1021) queries, ensuring a comprehensive range of higher-order reasoning tasks.

%% file: subtex/experiments.tex
\section{Benchmark Experiments}
We run two tasks, including dataset retrieval and open-domain question answering on our dataset, and benchmark mainstream models here.
\vspace{-10pt}
\subsection{Task Formulation}
For the dataset retrieval task, the system is given a question $q$ and should find the dataset and its attached information that is helpful to answer $q$ from the given dataset collection $\mathcal{C}$. The retrieval performance is measured using $R@k, k=\{1,5,20,100\}$ which means the recall rate of the positive dataset among $top-k$ retrieved results, and $MRR@100$ which means the mean reciprocal rank of the ground-truth dataset.

For the open-domain question answering (QA) task, the system is given a question $q$ and should generate a free-form answer responding to $q$. Unlike multi-choice QA tasks which are easily evaluated with exact-match based accuracy, free-form QA lacks a clear and reliable metric for evaluating the quality of answers, especially for long-form answers. As demonstrated in recent studies~\cite{evouna:2018}, leveraging entailment models (which determine whether a hypothesis can be logically inferred from a given premise) derived from natural language inference tasks can more effectively capture semantic similarity and achieve superior human alignment scores, surpassing even robust LLM-as-a-judge like GPT-3.5. Thus, in this work, we use an entailment model to measure the degree of response alignment with a reference answer. Following previous work, if the score given by the entailment model is larger than 0.5, then the answer is regarded as semantically equivalent to the reference answer. Then accuracy will be calculated as normal. To align with current long-form QA conventions, we also report ROUGE-L for the categories of long-answer types.
\begin{table*}[h]
\centering
\begin{tabular}{c|ccccc|ccccc}
\toprule
\multicolumn{1}{c|}{} & \multicolumn{5}{c|}{\textbf{w/o paper}} & \multicolumn{5}{c}{\textbf{w/ paper}} \\
\midrule
\textbf{Method} & \textbf{R@1} & \textbf{R@5} & \textbf{R@20} & \textbf{R@100} & \textbf{MRR@100} & \textbf{R@1} & \textbf{R@5} & \textbf{R@20} & \textbf{R@100} & \textbf{MRR@100} \\
\midrule
\textbf{BM25}  & 0.241 & 0.360 & 0.460 & 0.577 & 0.300 & 0.277 & 0.437 & 0.550 & \textbf{0.769} & \textbf{0.357} \\
\midrule
\textbf{DPR} & 0.060 & 0.119 & 0.218 & 0.375 & 0.095 & 0.061 & 0.130 & 0.241 & 0.415 & 0.100 \\
\textbf{Contriever} & 0.213 & 0.338 & 0.433 & 0.565 & 0.272 & 0.200 & 0.334 & 0.464 & 0.630 & 0.266 \\
\textbf{Specter2} & 0.172 & 0.320 & 0.435 & 0.570 & 0.244 & 0.173 & 0.324 & 0.448 & 0.600 & 0.248 \\
\textbf{E5-base-v2} & 0.202 & 0.336 & 0.468 & 0.592 & 0.269 & 0.224 & 0.377 & 0.518 & 0.674 & 0.299 \\
\textbf{E5-large-v2} & 0.202 & 0.336 & 0.468 & 0.592 & 0.269 & 0.224 & 0.377 & 0.518 & 0.674 & 0.299 \\
\textbf{BGE-m3} & 0.235 & 0.368 & 0.476 & 0.607 & 0.300 & 0.268 & 0.428 & \textbf{0.553} & 0.742 & 0.346 \\
\textbf{NV-Embed-v2} & \textbf{0.274} & \textbf{0.420} & \textbf{0.522} & \textbf{0.632} & \textbf{0.342} & \textbf{0.287} & \textbf{0.440} & 0.529 & 0.653 & \textbf{0.357} \\

\bottomrule
\end{tabular}
\caption{Performance of eight sparse and dense retrieval baselines on ScIRGen-Geo. Recall@{1, 5, 20, 100} and MRR@100 are reported under two index configurations—metadata-only (w/o paper) versus metadata + parsed-paper passages (w/ paper).}
\label{tab:4retrieval_bench}
\vspace{-20pt}
\end{table*}
\subsection{Benchmark Experiments for Dataset Retrieval}

In the dataset retrieval task, the dataset collection $\mathcal{C}$ consists of 3354 datasets as shown in Table~\ref{tab:dataset_stats}, partitioned into train/val/test sets at the dataset level (80:15:5 ratio). Due to QA density variation across datasets, the split yielded 49,200 (79.6\%), 8,784 (14.2\%), and 3,392 (5.5\%) QA pairs respectively. We use the questions and datasets in the test set only to benchmark the performance of retrievers. For retrievers, we choose representative retrievers in this field including sparse retrieval (BM25), dense retrieval across different sizes including general llm-based retrieval (DPR, Contriever~\cite{contriever:2021}, BGE-M3\cite{bgem3:2024}, E5~\cite{e5:2022}, etc.), and scientific retrieval (SPECTER \cite{specter:2020}) models.

We run retrieval on two different settings to assess the utility of our paper-augmented dataset representation. The w/o paper setting uses only the dataset title and dataset metadata in index construction, while the w/ paper setting also leverages relevant structured information obtained from the parsing phase of our data generation pipeline to enhance the index.

The results are shown in Table~\ref{tab:4retrieval_bench}. All the retrievers are evaluated without post-training in our dataset. We noticed that without adapting to a specific domain, some dense retriever models without large-scale and diverse pertaining like DPR fall behind simple sparse baselines like BM25, which aligns with the findings from BEIR~\cite{beir:2021} that raise concern about the generalizability of dense retrieval models. Notably, in the w/ paper setting BM25 is even comparable to SOTA llm-based retrieval methods. This phenomenon might be due to the characteristic of multi-aspect information and the widespread usage of professional terminology, which benefits keyword matching more than an all-in-one dense embedding.

Comparing the two dataset representation settings, all models perform significantly better with information from the paper. This improvement is the most significant in long-context embedding models like bge-m3 that can fully leverage the augmented information, which has a maximum of 8192 allowed input tokens, and has improved by 14\% in recall@100 by incorporating papers. 

Comparing different BERT-initialized models, we find that large pretrained models perform better than fine-tuned retrievers, demonstrating greater robustness in out-of-domain retrieval tasks. Scientific domain fine-tuned models like Specter2 are better than DPR, which is only fine-tuned on simple QA datasets. However, since our dataset focuses on a research-level and specific domain, the improvement brought by scientific fine-tuning still seems to lag behind the large-scale retriever pre-training. 

\begin{table}[h]
\centering
\scriptsize
\begin{tabular}{@{}ll|llllll@{}}
\toprule
\multicolumn{1}{l|}{\multirow{2}{*}{}} & \multirow{2}{*}{\textbf{avg.}} & \multicolumn{6}{c}{\textbf{Cognitive Level}} \\ \cmidrule(l){3-8} 
\multicolumn{1}{l|}{} &  & \textbf{C1} & \textbf{C2} & \textbf{C3} & \textbf{C4} & \textbf{C5} & \textbf{C6} \\\midrule
\multicolumn{1}{l|}{\textbf{Llama-2-7b-chat-hf}} & 0.15 & 0.15 & 0.13 & 0.12 & 0.13 & 0.27 & 0.13 \\
\multicolumn{1}{l|}{\textbf{Llama-2-13b-chat-hf}} & 0.21 & 0.25 & 0.16 & 0.25 & 0.19 & 0.27 & 0.26 \\
\multicolumn{1}{l|}{\textbf{Qwen-2.5-1.5B-Instruct}} & 0.16 & 0.16 & 0.14 & 0.17 & 0.14 & 0.26 & 0.16 \\
\multicolumn{1}{l|}{\textbf{Qwen-2.5-32B-Instruct}} & 0.21 & 0.21 & 0.24 & 0.12 & 0.17 & 0.32 & 0.16 \\
\multicolumn{1}{l|}{\textbf{DeepSeek-R1-Distill-Qwen-32B}} & 0.18 & 0.20 & 0.15 & 0.12 & 0.14 & 0.31 & 0.16 \\\midrule
\multicolumn{1}{l|}{\textbf{Galactica-1.3b}} & 0.28 & 0.25 & 0.25 & 0.27 & 0.27 & 0.43 & 0.25 \\
\multicolumn{1}{l|}{\textbf{Galactica-6.7b}} & 0.28 & 0.25 & 0.26 & 0.28 & 0.25 & 0.41 & 0.25 \\
\multicolumn{1}{l|}{\textbf{K2}} & 0.20 & 0.20 & 0.15 & 0.15 & 0.19 & 0.27 & 0.28 \\
\multicolumn{1}{l|}{\textbf{SciGLM-6B}} & 0.26 & 0.22 & 0.29 & 0.17 & 0.23 & 0.42 & 0.25 \\\midrule
\multicolumn{1}{l|}{\textbf{GPT-3.5-Turbo}} & 0.18 & 0.22 & 0.18 & 0.10 & 0.13 & 0.29 & 0.12 \\
\multicolumn{1}{l|}{\textbf{GPT-4}} & 0.17 & 0.21 & 0.18 & 0.10 & 0.12 & 0.27 & 0.12 \\\bottomrule
\end{tabular}
\caption{Cognitive-level QA benchmark for twelve LLMs. For each model, accuracy is averaged within the six Revised Bloom tiers (C1–C6).}
\label{tab:bench_coglevel}
\vspace{-25pt}
\end{table}

\begin{table}[h]
{\fontsize{9}{12}\selectfont
\begin{tabular}{@{}l|c|c|cc@{}}
\toprule
\multicolumn{1}{c|}{\multirow{2}{*}{}} & \multirow{2}{*}{Avg.} & \textit{Short-form} & \multicolumn{2}{c}{\textit{Long-form}} \\ \cmidrule(l){3-5} 
\multicolumn{1}{c|}{} & & Accuracy & Accuracy & ROUGE-L \\ \midrule
\textit{\textbf{Qwen2.5-1.5B}} & 0.161 & 0.265  & 0.158 & 0.222\\\cdashline{1-5}
\textbf{Qwen2.5-1.5B$_{+k=1}$} & 0.274 & 0.429 & 0.244 & 0.258\\\cdashline{1-5}
\textbf{Qwen2.5-1.5B$_{+k=5}$} & 0.274 & 0.476 & 0.230 & 0.270 \\\midrule
\textbf{Llama2-7B} & 0.153 & 0.273 & 0.149 & 0.255 \\\cdashline{1-5}
\textbf{Llama2-7B$_{+k=1}$} & 0.306 & 0.490 & 0.263 & 0.273\\\cdashline{1-5}
\textbf{Llama2-7B$_{+k=5}$} & 0.324 & 0.520 & 0.269 & 0.307 \\\midrule
\textbf{SciGLM-6B} & 0.263 & 0.302 &  0.267 & 0.247 \\\cdashline{1-5}
\textbf{SciGLM-6B$_{+k=1}$} & 0.269 & 0.389 & 0.246 & 0.178\\\cdashline{1-5}
\textbf{SciGLM-6B$_{+k=5}$} & 0.229 & 0.373 & 0.197 & 0.192 \\\midrule
\textbf{K2} & 0.198  & 0.335 & 0.180 & 0.284\\\cdashline{1-5}
\textbf{K2$_{+k=1}$} & 0.312 & 0.517 & 0.256 & 0.303 \\\cdashline{1-5}
\textbf{K2$_{+k=5}$} & 0.338 & 0.556 & 0.278 & 0.319 \\\bottomrule
\end{tabular}
\caption{Effect of retrieval-augmented prompting on QA performance. Accuracy (short/long) and ROUGE-L are reported for four LLMs with no retrieval, top-1 passage, and top-5 passages retrieved by BGE-m3.}
\label{tab:rag}
}
\vspace{-30pt}
\end{table}

\begin{table*}[h]
{\fontsize{6.5}{11}\selectfont
\begin{tabular}{@{}l|llllllllllllllllll@{}}
\toprule
\multirow{2}{*}{} & \multicolumn{18}{c}{\textbf{Question Category}} \\ \cmidrule(l){2-19} 
                  & Ver. & Disj. & Conc. & Feat. & Quan. & Def. & Exam. & Comp. & Int. & Cas.A. & Cas.C. & Goal. & Proc. & Enab. & Exp. & Judg. & As. & Req. \\
\midrule
\textbf{Llama-2-7b}    & 0.861 & 0.529 & 0.106 & 0.152 & 0.069 & 0.245 & 0.033 & 0.166 & 0.086 & 0.071 & 0.139 & 0.130 & 0.077 & 0.061 & 0.096 & 0.149 & 0.063 & 0.270 \\
\textbf{Llama-2-13b}   & 0.722 & 0.549 & 0.253 & 0.491 & 0.085 & 0.278 & 0.047 & 0.225 & 0.218 & 0.117 & 0.148 & 0.111 & 0.206 & 0.071 & 0.084 & 0.220 & 0.114 & 0.336 \\
\textbf{Qwen-2.5-1.5b}  & 0.765 & 0.637 & 0.126 & 0.116 & 0.081 & 0.231 & 0.079 & 0.179 & 0.201 & 0.079 & 0.177 & 0.102 & 0.145 & 0.086 & 0.079 & 0.099 & 0.103 & 0.125 \\
\textbf{Qwen-2.5-32b}   & 0.826 & 0.706 & 0.126 & 0.094 & 0.126 & 0.429 & 0.159 & 0.325 & 0.236 & 0.142 & 0.215 & 0.255 & 0.089 & 0.046 & 0.146 & 0.149 & 0.154 & 0.112 \\
\textbf{DeepSeek-R1-32b}& 0.861 & 0.775 & 0.136 & 0.085 & 0.130 & 0.245 & 0.084 & 0.225 & 0.172 & 0.088 & 0.196 & 0.134 & 0.097 & 0.051 & 0.112 & 0.149 & 0.126 & 0.112\\
\midrule
\textbf{Galactica-1.3b}  & 0.861 & 0.618 & 0.242 & 0.286 & 0.102 & 0.439 & 0.065 & 0.219 & 0.431 & 0.146 & 0.464 & 0.259 & 0.218 & 0.279 & 0.270 & 0.270 & 0.229 & 0.099 \\
\textbf{Galactica-6.7b}  & 0.861 & 0.716 & 0.187 & 0.344 & 0.130 & 0.396 & 0.173 & 0.205 & 0.379 & 0.133 & 0.421 & 0.176 & 0.222 & 0.259 & 0.191 & 0.390 & 0.211 & 0.105 \\
\textbf{SciGLM}         & 0.817 & 0.647 & 0.177 & 0.196 & 0.089 & 0.443 & 0.126 & 0.291 & 0.356 & 0.117 & 0.493 & 0.352 & 0.149 & 0.157 & 0.118 & 0.220 & 0.240 & 0.250 \\
\textbf{K2}             & 0.878 & 0.539 & 0.192 & 0.317 & 0.102 & 0.226 & 0.047 & 0.179 & 0.253 & 0.096 & 0.129 & 0.088 & 0.109 & 0.086 & 0.107 & 0.220 & 0.131 & 0.441 \\
\midrule
\textbf{GPT-3.5}        & 0.939 & 0.716 & 0.126 & 0.116 & 0.150 & 0.321 & 0.131 & 0.232 & 0.132 & 0.083 & 0.129 & 0.167 & 0.069 & 0.030 & 0.101 & 0.085 & 0.063 & 0.184 \\
\textbf{GPT-4}          & 0.904 & 0.676 & 0.152 & 0.125 & 0.122 & 0.311 & 0.126 & 0.285 & 0.126 & 0.050 & 0.153 & 0.144 & 0.040 & 0.061 & 0.056 & 0.078 & 0.051 & 0.164 \\
\bottomrule
\end{tabular}
}
\caption{Question-type QA benchmark that evaluates different LLMs across all 18 types in Graesser and Pearson’s taxonomy}
\label{tab:bench_cate}
\vspace{-20pt}
\end{table*}

\subsection{Benchmark Experiments for QA Tasks}

In the QA benchmark, we test LLMs on our ScIRGen-Geo Dataset. We choose three types of representative LLMs, including general open-source LLMs (LLama-2, Qwen-2.5, DeepSeek-R1), scientific foundation LLMs (Galactica, K2, SciGLM), and closed-source LLMs (GPT). We also cover different parameter sizes from 1.3B to 32B in the chosen LLMs. The models are asked to answer the question briefly.

We studied the QA performance across different question categories and different cognitive levels according to Bloom's taxonomy, as shown in Table~\ref{tab:bench_coglevel}. In all, we find that scientific foundation models perform the best among all the models, with Galactica-6.7B performing the best. Larger models generally perform better, especially in categories that are reasoning-intensive, such as Quantification and Causal inference. For different cognitive levels, we find that C3(Applying) and C4(Evaluating) are the hardest types, and C5(Evaluating) is the level that achieves the highest accuracy. While difficulty plays an important role here, we find that the ambiguity in evaluating C5 and C6(Creating) level questions may also influence the scoring. The Table ~\ref{tab:bench_cate} is the QA benchmark classified by different question categories. We find that verification questions (Ver.) consistently achieve high scores across models, with GPT models standing out, demonstrating that large language models are particularly adept at handling verification tasks. Second, reasoning-intensive qa types like causal inference generally yield uniformly low scores (generally below 0.25) across all models, signaling a common challenge in addressing this question type. 

We also experiment on whether incorporating retrieval enhances the question answering performance. We chunk the passages into 100 tokens per passage, which results in 25145 passages. We then apply BGE-m3 as the retriever and feed the top-1 and top-5 retrieved passages to the generator models along with the question in the prompt. As shown in Table~\ref{tab:rag}, using extra retrieved content largely improves QA performance, and for small k in our setting, more context will bring better performance. An exception is SciGLM, which we suspect is a result of losing some long-context comprehension ability during scientific-specific finetuning.

%% file: subtex/relatedwork.tex
\vspace{-15pt}
\section{Related Work}
\paragraph{Dataset Generation.} Constructing high-quality datasets for complex QA and retrieval tasks remains a challenge. Traditional methods, such as crowdsourcing, can be expensive, especially in expert domains, while automatic approaches to construct multi-hop questions by chaining simple questions~\cite{hotpotqa:2018, musique:2022} are limited in diversity. Data collection from community forums~\cite{bright:2025} can generate diverse and realistic data, but often suffers from noise and is restricted to active online domains. More recently, PLMs have been explored for dataset generation by leveraging their massive knowledge from large-scale pretraining~\cite {plm4nlg:2019}. Researchers fine-tune PLMs with human-annotated seed set~\cite{filter_roundtrip:2019, trainqa:2020, domainqg:2021}, use in-context prompting~\cite{alpaca:2023,selfinstruct:2023} and design agent cooperation workflows~\cite{agentinstruct:2024, arena_learning:2024} to enable PLMs for question and answer generation. However, the lack of controllability and verifiability of LM-generated content hampers the utility of synthetic data. As a result, there still remains a significant shortage of useful RAG datasets, particularly in expert domains such as scientific research.

\vspace{-10pt}
\paragraph{Scientific Retrieval and QA}

Recent advances in scientific AI systems focus on two interconnected directions: foundation model-powered QA systems and scientific resource retrieval. While scientific foundation models~\cite{galactica:2022, k2:2024} exhibit broad scientific capabilities, their professional reliability is limited by hallucination risks from misaligned knowledge boundaries and untraceable source grounding~\cite{specter:2020, alce:2023}. Some seek to address credibility through retrieval, which has been formulated as literature retrieval, citation retrieval and dataset retrieval. Literature retrieval and citation retrieval~\cite{citerec:2010, scibert:2019, litsearch:2024, dorismae:2023} aim to find relevant papers given a query or related content, whereas dataset recommenders~\cite{datasetrecom:2021, datafinder:2023} attempt to link researchers with concrete experimental resources. However, current scientific RAG systems are still short in understanding realistic and professional queries and identifying heterogeneous scientific resources at scale~\cite{bright:2025}, impeding sophisticated and informed decision-making in scientific discovery. Notably, scientific workflows inherently demand multi-stage reasoning, as task-critical questions arise throughout problem-solving, requiring robust modeling frameworks and optimization pipelines~\cite{lyu2024autostf, wang2025language, wang2025phyda, guo2025revisiting}. This complexity highlights the urgent need for RAG systems that can handle multi-stage reasoning and cross-resource synthesis in authentic scientific contexts.


%% file: subtex/limitation_conclusion.tex
\vspace{-10pt}
\section{Conclusion}
In conclusion, ScIRGen provides a systematic, end-to-end framework for constructing realistic retrieval and question answering datasets tailored to the demands of scientific research. By integrating an information extraction pipeline to encompass both datasets and relevant scholarly literature, along with leveraging Graesser’s comprehensive question taxonomy, and incorporating a robust adaptation of the $\Delta \fancyname$ framework to rigorously verify the fidelity of free-form answers, our approach ensures that generated queries reflect the depth and complexity of actual research inquiries. The resulting ScIRGen-Geo dataset underscores the multifaceted nature of scientific workflows, revealing that contemporary models still face significant challenges in retrieval and reasoning for domain-specific, task-oriented questions. Beyond its immediate contribution to geoscience, the ScIRGen framework further offers a generalizable pipeline to synthesize data to catalyze the development of advanced knowledge discovery systems across diverse scientific fields.

\vspace{-10pt}
\section{Acknowledgement}

This work was supported in part by the National Key R\&D Program of China (Grant No. 2023YFF0725001), in part by the National Natural Science Foundation of China (Grant No.92370204), in part by the Guangdong Basic and Applied Basic Research Foundation (Grant No. 2023B1515120057), in part by The Education Bureau of Guangzhou Municipality, in part by the Guangzhou Basic and Applied Basic Research Program (Grant No. 2024A04J3279).

%% file: subtex/prompt_appendix.tex
\section{Appendix}


\label{appendix:paper_extraction_prompt}
\begin{figure*}[h]
    \centering
    \includegraphics[width=\textwidth]{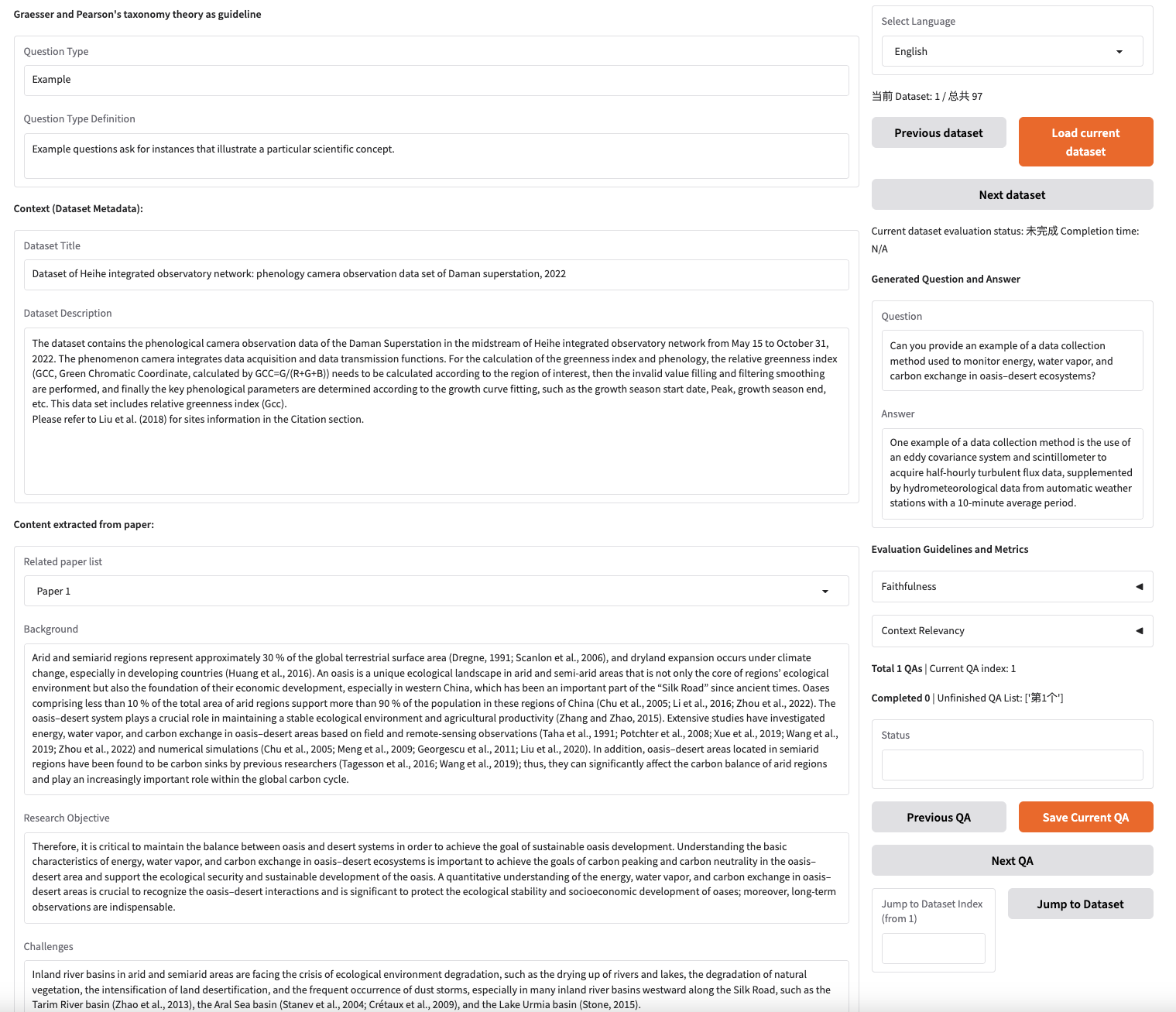}
    \caption{Expert Annotation Interface.}
    \label{fig:annotation}
\end{figure*}

Table \ref{tab:model_comparison} presents the retriever parameters used in our retrieval benchmark. Tables \ref{prompt:corpus_curation-a}, \ref{prompt:corpus_curation-b}, \ref{prompt:corpus_curation-c}, and \ref{prompt:corpus_curation-d} detail the prompts used in the Corpus Curation stage. Tables \ref{prompt:main_generation}, \ref{prompt:generate_for_without_pdf} detail the prompts used in the QA data generation stage. Table \ref{prompt:cogntive_level} details the prompts employed during the Evaluation stage. Figure \ref{fig:annotation} displays the interface utilized by experts to annotate data.

\begin{table}[h]
\centering
\begin{tabular}{lccc}
\toprule
\textbf{Method} & \textbf{Parameters} & \textbf{Max Input Length}\\
\midrule
BM25 & - & - \\
DPR & 110M & 512  \\
Contriever & 110M & 512  \\
Specter2 & 110M & 512 \\
E5-base-v2 & 110M & 512 \\
E5-large-v2 & 336M & 512  \\
BGE-m3 & 550M & 8192  \\
NV-Embed-v2 & 7B & 32768 \\
\bottomrule
\end{tabular}
\caption{Comparison of retrieval methods on ScIRGen-Geo with model parameters and maximum input length.}
\label{tab:model_comparison}
\end{table}

\begin{table*}[h]
\centering

\begin{tabular}{p{\linewidth}}
\toprule
Analyze the academic paper and determine if it likely uses a dataset with characteristics similar to:

Target Dataset Information: {}

Instructions:

1. Carefully read the paper content and look for any direct mentions of the target dataset.

2. Also consider indirect references, such as descriptions of a dataset with characteristics closely matching the target dataset.

3. If there are no direct mentions but strong indirect evidence, err on the side of considering the paper relevant.

4. Return your response in the following format:

USED:[Yes/No]

EXPLANATION: [Brief explanation of your decision with specific evidence from the text]\\
\bottomrule
\end{tabular}
\caption{(a) Dataset Relevance Analyzer}
\label{prompt:corpus_curation-a}
\end{table*}

\begin{table*}[h]
\centering
\resizebox{\linewidth}{!}{%
\begin{tabular}{p{\linewidth}}
\toprule
You are an expert in academic paper analysis. Given a segment of text from an academic paper, classify it into one of these categories:

1. abstract\&introduction

2. related works

3. method

4. experiment

5. conclusion

Rules:
- If the text doesn't clearly belong to any category, classify it as "none"

- The abstract and introduction sections should be combined into "abstract\&introduction"

- Method sections might also be called "methodology", "approach", "proposed method", etc.

- Experiment sections might also be called "results", "evaluation", "experimental results", etc.

- Related works might also be called "background", "literature review", etc.

Return ONLY the category name, nothing else.\\

\bottomrule
\end{tabular}
}
\caption{(b) Paper Segment Classifier}
\label{prompt:corpus_curation-b}
\end{table*}

\begin{table*}[h]
\centering
\resizebox{\linewidth}{!}{%
\begin{tabular}{p{\linewidth}}
\toprule
You are an experienced researcher. Analyze the text and extract COMPLETE SENTENCES that contain research components based on these specific definitions:

1. Background:

- Provide complete sentences that explain essential background knowledge related to the research topic.

- Describe the current state of knowledge in the field of study, highlighting key findings and challenges.

- Include explanations of key concepts, theories, or phenomena that are directly relevant to the research.

Requirements:

- The background must be directly related to the paper's research topic.

- It should offer context necessary for understanding the research and its significance.

- The background should explain relevant theories, concepts, or phenomena that lay the groundwork for the study.

2. Research Objective:

- Provide complete sentences that state the primary purpose of the research, or,

- Describe the main goal or target of the study and the specific problem the research seeks to solve or address, or,

- Explain what the researchers hope to achieve through their study, such as providing a solution, filling a gap in knowledge, testing a hypothesis, or understanding a phenomenon, or,

- Clarify the scope of the research, whether it's a theoretical exploration, a practical application, or a combination of both, or,

- Outline any specific hypotheses or questions the study aims to test or answer.

Requirements:

- State the research motivation, explaining why this study is important and what drives the inquiry.

- Specify the research goals, outlining the desired outcomes and what the researchers hope to uncover or demonstrate.

- Describe what the research seeks to achieve or investigate, though flexibility in expression is allowed depending on the context of the study.

- It may also help to provide context for how the research fits into existing literature or contributes to the field, although this is not always necessary.

3. Methods (ONLY include if present):

- ONLY complete sentences describing experimental procedures ACTUALLY PERFORMED in THIS paper

- ONLY complete sentences explaining methodological steps ACTUALLY IMPLEMENTED by THIS study's authors

Requirements:

- Include ONLY methods that were directly executed in THIS research

- Must be preceded by phrases like "we", "our study", "in this paper", or similar indicators showing the authors performed these methods
\\
\bottomrule
\end{tabular}
}
\label{prompt:corpus_curation-c}
\end{table*}

\begin{table*}[h]
\centering
\resizebox{\linewidth}{!}{%
\begin{tabular}{p{\linewidth}}
\toprule

- EXCLUDE ALL sentences that:

    * Describe other researchers' methods
    
    * Review existing approaches
    
    * Discuss general methodological concepts
    
    * Start with phrases like "Previous studies...", "Existing methods...", "Other researchers..."
    
    * Appear in literature review or related work sections
    
- Each extracted method must describe a specific action or procedure performed by the current study's authors

4. Challenges (ONLY include if present):

- Complete sentences describing research gaps or inadequacies in current solutions

- Complete sentences explaining limitations or difficulties in the research process

- Complete sentences identifying technical barriers or obstacles

- Complete sentences describing unresolved problems in the field

Requirements: Extract full sentences that clearly describe specific challenges, limitations, or difficulties faced during the research.

5. Dataset (ONLY include if present):

- Complete sentences describing the datasets used

- Complete sentences detailing size, composition, or key characteristics
Requirements: 

- Extract full sentences that provide complete information about the dataset

- Do not split dataset descriptions mid-sentence

- Only include datasets actually used in THIS study's experiments

6. Findings (ONLY include if present):

- ONLY complete sentences stating PRIMARY research findings that directly address the core research question

- ONLY complete sentences describing MAJOR discoveries or conclusions that answer the study's main objectives

Requirements: 

- Each finding must directly correspond to the paper's core research questions/objectives

- Include ONLY high-level, significant findings that represent the paper's main contributions

- EXCLUDE:

    * Intermediate experimental results
    
    * Detailed performance metrics from individual experiments
    
    * Secondary or supporting findings
    
    * Preliminary results
    
    * Ablation study results
    
    * Comparative experiment outcomes unless they represent the paper's main conclusion
    
Example of what to include:

    * Major methodological breakthroughs
    
    * Primary theoretical contributions
    
    * Key empirical discoveries that advance the field
    
    * Central conclusions that answer the research question

Instructions:

1. ONLY extract COMPLETE SENTENCES. Never extract partial sentences or fragments.

2. Each extracted text must be grammatically complete and semantically self-contained.

3. If multiple sentences are needed to express a complete idea, include all relevant sentences together.

4. If a sentence spans multiple categories, include it in all relevant categories.

5. Maintain the exact wording from the original text - do not modify or summarize.

6. If no complete sentences are found for a category, write "None" for that category.

Format your response as key-value pairs, one per line:

Background: [extracted complete sentences]
Research Objective: [extracted complete sentences]
Methods: [extracted complete sentences]
Challenges: [extracted complete sentences]
Dataset: [extracted complete sentences]
Findings: [extracted complete sentences]
Example of good extraction:

Original text: "Previous studies have used CNN-based approaches for image classification. Several researchers implemented attention mechanisms with varying success. In this paper, we propose a novel hybrid architecture combining transformers with CNNs.  We trained our model using a three-stage process with curriculum learning. Our experiments showed 97\% accuracy on the test set."

Background: Previous studies have used CNN-based approaches for image classification. Several researchers implemented attention mechanisms with varying success.

Research Objective: In this paper, we propose a novel hybrid architecture combining transformers with CNNs.

Methods: We trained our model using a three-stage process with curriculum learning.

Findings: Our experiments showed 97\% accuracy on the test set.

Challenges: None

Dataset: None

IMPORTANT:

Please think twice before you response, make sure you understand and confirm your answer.

Find as many sentences as you can.

The more content you find, the better. Do not miss any possible content.
\\
\bottomrule
\end{tabular}
}
\caption{(c) Fine-grained Paper Parse Engine-1.}
\label{prompt:corpus_curation-c}
\end{table*}

\begin{table*}[h]
\centering
\resizebox{\linewidth}{!}{

\begin{tabular}{p{\linewidth}}
\toprule
You are an expert research analyst. Evaluate the quality and relevance of the extracted research content. 

Evaluation Requirements:
1. EACH section (Background, Methods, Findings, Challenges, Dataset) MUST have AT LEAST ONE item retained

2. For sections with multiple items:

- Keep the specific and informative items

- Filter out redundant content

Content Selection Principles:

1. Completeness: Content should form a complete, self-contained unit of information

2. Relevance: Content should directly relate to its category

3. Non-redundancy: Avoid repetitive information

4. Clarity: Prefer clear, well-structured expressions

5. Category Alignment: Content should strictly belong to its claimed category

Detailed Evaluation Guidelines:

1. Each category must maintain at least one item

2. Selected content should be complete and self-contained

3. Avoid splitting related information across multiple items

4. Maintain proper categorization of content

5. Prefer comprehensive items over partial descriptions

6. Remove redundant information even if well-written

7. Ensure selected content actually belongs to its category

8. Consider the overall coherence of the selected content

Here's a concrete example of good evaluation:

Original Content Example (partial):

Background:

1. "Energy and water vapor interactions between land surfaces and the 
atmosphere are the most crucial ecological processes..." [Complete EC and LAS background]

2. "Experimental Research (W A TER)", many observation sites were established..."

3. "The relationship between $H + LE$ and $R_n$-$G_0$ can be expressed by..."

Analysis and Decisions:

- Keep item 1 because it provides comprehensive background about methods and limitations

- Remove item 2 because it's an incomplete sentence lacking context

- Remove item 3 because it contains results discussion rather than background

For your evaluation task:

1. Analyze each section's content carefully

2. Apply the selection principles consistently

3. Provide clear reasoning for your decisions

4. Maintain the essential meaning while removing redundancy

5. Ensure proper categorization is maintained

Return your response in this format:

KEEP-INDICES:

Background: [list of indices to keep]
Methods: [list of indices to keep]
Findings: [list of indices to keep]
Challenges: [list of indices to keep]
Dataset: [list of indices to keep]
REASON: [Brief explanation of your decisions, with reference to specific principles applied]

Remember:
- DO NOT modify any of the extracted content

- Keep at least one item per section if any exists

- Focus on quality over quantity while ensuring coverage of key information

- Consider how selected items work together to tell a complete story\\
\bottomrule
\end{tabular}
}
\caption{(c) Fine-grained Paper Parse Engine-2}
\label{prompt:corpus_curation-d}
\end{table*}


\begin{table*}[h]
\centering
\resizebox{\linewidth}{!}{%
\begin{tabular}{p{\linewidth}}
\toprule

You are a researcher generating questions and answers to find relevant metadata, datasets, and papers within a specific domain.

Below are the potential question types. Choose the type that best fits the field information and the user's purpose.

1. \textbf{Verification}: Verification questions seek a simple 'yes' or 'no' answer to confirm specific details.

2. \textbf{Disjunctive}: Disjunctive questions present multiple options, asking the researcher to identify which one is applicable.

\textbf{...}  The other Expert-Elicited Definitions for each question type are provided in Table \ref{tab:graesser-scheme}. 


Task: Based on the following dataset metadata, return the most appropriate 8 question types. Give me the name of each type and not other information.

\textbf{Dataset Metadata:} \{dataset title and description\}
\\
\bottomrule
\end{tabular}
}
\caption{Question Type Selection Prompt}
\label{prompt:generate_for_without_pdf}
\end{table*}

\begin{table*}[h]
\centering
\resizebox{\linewidth}{!}{%
\begin{tabular}{p{\linewidth}}
\toprule
\textbf{SYSTEM ROLE:} 
You are a researcher asking questions aiming to find relevant metadata, datasets, and papers within a specific domain.

\textbf{USER ROLE:}

I am exploring the field above. I'm interested in finding datasets or understanding data collection methods related to this field.

\textbf{Field:}

- \textbf{Metadata}: \{dataset title and description\}

- \textbf{Content of relevant Papers}: \{detail content\}

Please generate three questions that aim to discover or explore general information about data collection techniques, challenges, and potential datasets in this domain under the question type below and answer each query using the field above.

- \textbf{Question Type}: \{detail type\} \\
- \textbf{Question Definition}: \{detail definition\} \\
- \textbf{Question Example}: \{detail example\}

\textbf{Guidelines:}

1. Only questions and answers without any other information.

2. Do not summarize the metadata or Content of relevant Papers.

3. Focus on questions that are open-ended and explore a wide range of possibilities or methodologies related to the field.

4. Use neutral terms like "a dataset," "data collection method," or "research approach," instead of specific references like "the study" or "this dataset."

5. Aim for questions that encourage exploration of nuanced technological details, methodological rigor, and potential sources or strategies for dataset expansion or refinement within the field.

6. The answer should come from the field I mentioned above.

7. Return the questions and answers in JSON format.\\
\bottomrule
\end{tabular}
}
\caption{Main Prompt to generate question and answer}
\label{prompt:main_generation}
\end{table*}

\label{appendix:evaluation_prompt}
\begin{table*}[h]
\centering
\resizebox{\linewidth}{!}{%
\begin{tabular}{p{\linewidth}}
\toprule
\textbf{Role:} You are an expert evaluator and cognitive scientist specializing in revised Bloom’s Taxonomy.

\textbf{Task:}Classify each question based on its cognitive complexity into one of the six categories (C1--C6) from the revised Bloom’s Taxonomy.
\textbf{CATEGORIES AND DESCRIPTIONS:}

\textbf{C1 (Remembering):}\\
- Retrieve relevant knowledge from long-term memory.\\
- Focus on recalling or recognizing facts, definitions, or basic concepts.\\
\textbf{Examples:}\\
-- RECOGNIZING (identifying).\\
-- RECALLING (retrieving).\\

\textbf{C2 (Understanding):}\\
- Construct meaning from information through interpretation, explanation, or comparison.\\
- Emphasis is on understanding concepts rather than simply recalling them.\\
\textbf{Examples:}\\
-- INTERPRETING (clarifying, paraphrasing, representing, translating).\\
-- EXEMPLIFYING (illustrating, instantiating).\\
-- CLASSIFYING (categorizing, subsuming).\\
-- SUMMARIZING (abstracting, generalizing).\\
-- INFERRING (concluding, extrapolating, predicting).\\
-- COMPARING (contrasting, mapping, matching).\\
-- EXPLAINING (constructing models).\\
\textbf{C3 (Applying):}\\
\bottomrule
\end{tabular}
}
\end{table*}

\begin{table*}[h]
\centering
\resizebox{\linewidth}{!}{%
\begin{tabular}{p{\linewidth}}
\toprule
- Use knowledge in a new context or apply learned procedures to solve problems.\\
- Emphasis is on performing tasks or carrying out processes.\\
\textbf{Examples:}\\
-- EXECUTING (carrying out).\\
-- IMPLEMENTING (using).\\

\textbf{C4 (Analyzing):}\\
- Break down information into parts to understand relationships, patterns, or structures.\\
- Focus on examining and organizing components.\\
\textbf{Examples:}\\
-- DIFFERENTIATING (discriminating, distinguishing, selecting).\\
-- ORGANIZING (finding coherence, outlining, parsing, structuring).\\
-- ATTRIBUTING (deconstructing).\\

\textbf{C5 (Evaluating):}\\
- Make judgments based on criteria, standards, or evidence.\\
- Focus on assessing or critiquing based on logical reasoning.\\
\textbf{Examples:}\\
-- CHECKING (coordinating, detecting, monitoring, testing).\\
-- CRITIQUING (judging).\\

\textbf{C6 (Creating):}\\
- Generate new ideas, designs, or constructs by combining elements in novel ways.\\
- Focus on producing original work or reassembling existing components into new patterns.\\
\textbf{Examples:}\\
-- GENERATING (hypothesizing).\\
-- PLANNING (designing).\\
-- PRODUCING (constructing).\\
\textbf{Classification Instructions}\\
1. \textbf{Identify the Key Action}\\
\quad - Examine the main verb(s) or action(s) in the single question (e.g., “compare,” “analyze,” “create,” “evaluate,” etc.).\\
2. \textbf{Match to Category}\\
\quad - Decide which of the six categories (C1--C6) the question’s primary cognitive demand aligns with.\\
\quad - \textbf{Tie-Breaking Rule:} If the question seems to fit multiple categories, select the one that best represents the dominant cognitive skill required.\\
3. \textbf{Focus on Cognitive Process, Not Subject Matter}\\
\quad - Base your classification on the mental operation needed (recalling, understanding, applying, analyzing, evaluating, or creating), rather than the topic itself.\\
4. \textbf{Output Format}\\
\quad - Provide \textbf{only} the category code and name in parentheses.\\
\quad - For instance: \texttt{C1(Remembering)}, \texttt{C2(Understanding)}, etc.\\
\quad - \textbf{No extra explanations or commentary}---just the classification code.\\

\textbf{Question to Classify:} \{question\}\\
\bottomrule
\end{tabular}
}
\caption{Cognitive Level Evaluation Prompt}
\label{prompt:cogntive_level}
\end{table*}

\label{appendix:evaluation_prompt}

